\documentclass[10pt,twocolumn,letterpaper]{article}

\usepackage{cvpr}
\usepackage{times}
\usepackage{epsfig}
\usepackage{graphicx}
\usepackage{amsmath}
\usepackage{amssymb}
\usepackage{array}
\usepackage{wrapfig}
\usepackage{booktabs}
\usepackage{multirow}
\usepackage{algorithm}
\usepackage{algorithmic}

\usepackage[breaklinks=true,bookmarks=false]{hyperref}

\cvprfinalcopy 

\def\cvprPaperID{****} 

\begin{document}

\title{Deep Multimodal Clustering for Unsupervised Audiovisual Learning\thanks{\copyright 2019 IEEE. Personal use of this material is permitted. Permission from IEEE must be obtained for all other uses, in any current or future media, including reprinting/republishing this material for advertising or promotional purposes, creating new collective works, for resale or redistribution to servers or lists, or reuse of any copyrighted component of this work in other works.}}

\author{Di Hu\\
Northwestern Polytechnical University\\
{\tt\small hdui831@mail.nwpu.edu.cn}
\and
Feiping Nie\\
Northwestern Polytechnical University\\
{\tt\small feipingnie@gmail.com}
\and
Xuelong Li\thanks{Corresponding author}\\
Northwestern Polytechnical University\\
{\tt\small xuelong\_li@ieee.org}
}

\maketitle

\begin{abstract}
The seen birds twitter, the running cars accompany with noise, etc. These naturally audiovisual correspondences provide the possibilities to explore and understand the outside world.
However, the mixed multiple objects and sounds make it intractable to perform efficient matching in the unconstrained environment.
To settle this problem, we propose to adequately excavate audio and visual components and perform elaborate correspondence learning among them.
Concretely, a novel unsupervised audiovisual learning model is proposed, named as \emph{Deep Multimodal Clustering} (DMC),
that synchronously performs sets of clustering with multimodal vectors of convolutional maps in different shared spaces for capturing multiple audiovisual correspondences.
And such integrated multimodal clustering network can be effectively trained with max-margin loss in the end-to-end fashion.
Amounts of experiments in feature evaluation and audiovisual tasks are performed.
The results demonstrate that DMC can learn effective unimodal representation, with which the classifier can even outperform human performance.
Further, DMC shows noticeable performance in sound localization, multisource detection, and audiovisual understanding.
\end{abstract}

\section{Introduction}
When seeing a dog, why the sound emerged in our mind is mostly barking instead of miaow or others? It seems easy to answer ``we can only catch the barking dog in our daily life''.
As specific visual appearance and acoustic signal usually occur together, we can realize that they are strongly correlated, which accordingly makes us recognize the sound-maker of visual dog and the distinctive barking sound.
Hence, the concurrent audiovisual message provides the possibilities to better explore and understand the outside world~\cite{holmes2005multisensory}.

The cognitive community has noticed such phenomenon in the last century and named it as multisensory processing~\cite{holmes2005multisensory}.
They found that some neural cells in superior temporal sulcus (a brain region in the temporal cortex) can simultaneously response to visual, auditory, and tactile signal~\cite{hikosaka1988polysensory}.
When the concurrent audiovisual message is perceived by the brain, such neural cells could provide corresponding mechanism to correlate these different messages, which is further reflected in various tasks, such as lip-reading~\cite{calvert1997activation} and sensory substitute~\cite{proulx2014multisensory}.

In view of the merits of audiovisual learning in human beings, it is highly expected to make machine possess similar ability, i.e., exploring and perceiving the world via the concurrent audiovisual message.
More importantly, in contrast to the expensive human-annotation, the audiovisual correspondence can also provide valuable supervision, and it is pervasive, reliable, and free~\cite{owens2016ambient}.
As a result, the audiovisual correspondence learning has been given more and more attention recently.
In the beginning, the coherent property of audiovisual signal is supposed to provide cross-modal supervision information, where the knowledge of one modality is transferred to supervise the other primitive one.
However, the learning capacity is obviously limited by the transferred knowledge and it is difficult to expand the correspondence into unexplored cases.
Instead of this, a natural question emerges out: can the model learn the audiovisual perception just by their correspondence without any prior knowledge?
The recent works give definitive answers~\cite{arandjelovic2017look,owens2018audio}.
They propose to train an audiovisual two-stream network by simply appending a correspondence judgement on the top layer.
In other words, the model learns to match the sound with the image that contains correct sound source.
Surprisingly, the visual and auditory subnets have learnt to response to specific object and sound after training the model, which can be then applied for unimodal classification, sound localization, etc.

The correspondence assumption behind previous works~\cite{owens2016ambient,arandjelovic2017look,aytar2017see} rely on specific audiovisual scenario where the sound-maker should exist in the captured visual appearance and single sound source condition is expected.
However, such rigorous scenario is not entirely suitable for the real-life video.
First, the unconstrained visual scene contains multiple objects which could be sound-makers or not, and the corresponding soundscape is a kind of multisource mixture.
Simply performing the global corresponding verification without having an insight into the complex scene components could result in the inefficient and inaccurate matching, which therefore need amounts of audiovisual pairs to achieve acceptable performance~\cite{arandjelovic2017look} but may still generate semantic irrelevant matching~\cite{senocak2018learning}.
Second, the sound-maker does not always produce distinctive sound, such as honking cars, barking dogs, so that the current video clip does not contain any sound but next one does, which therefore creates inconsistent conditions for the correspondence assumption.
Moreover, the sound-maker may be even out of the screen so we cannot see it in the video, e.g., the voiceover of photographer.
The above intricate audiovisual conditions make it extremely difficult to analyze and understand the realistic environment, especially to correctly match different sound-makers and their produced sounds.
So, a kind of elaborate correspondence learning is expected.

As each modality involves multiple concrete components in the unconstrained scene, it is difficult to correlate the real audiovisual pairs.
To settle this problem, we propose to disentangle each modality into a set of distinct components instead of the conventional indiscriminate fashion.
Then, we aim to learn the correspondence between these distributed representations of different modalities.
More specifically, we argue that the activation vectors across convolution maps have distinct responses for different input components, which just meets the clustering assumption.
Hence, we introduce the Kmeans into the two-stream audiovisual network to distinguish concrete objects or sounds captured by video.
To align the sound and its corresponding producer, sets of shared spaces for audiovisual pairs are effectively learnt by minimizing the associated triplet loss.
As the clustering module is embedded into the multimodal network, the proposed model is named as \emph{Deep Multimodal Clustering} (DMC). Extensive experiments conducted on wild audiovisual pairs show superiority of our model on unimodal features generation, image/acoustic classification and some audiovisual tasks, such as single sound localization and multisource {Sound Event Detection} (SED). And the ultimate audiovisual understanding seems to have preliminary perception ability in real-life scene.


\section{Related Works}
\begin{table}[h]
\begin{center}
  \newcolumntype{C}[1]{>{\centering}p{#1}}
\begin{tabular}{C{0.8cm}C{1.2cm}C{2.8cm}C{2cm}}
\toprule[2pt]
  Source  &  Supervis. & Task & Reference
\tabularnewline\midrule[1pt]
Sound                                           & Vision                & Acoustic Classif.   & \cite{aytar2016soundnet,harwath2016unsupervised,harwath2017learning,aytar2017see}  \tabularnewline  \hline
Vision                                           & Sound                 & Image Classif.        &\cite{owens2016ambient,aytar2017see}  \tabularnewline  \hline
Sound  &  \multirow{3}*{Match}  &  Classification           &   \cite{arandjelovic2017look}                  \tabularnewline
\&                                      &                                    &   Sound Localization                                      &\cite{arandjelovic2017objects,senocak2018learning,owens2018audio,zhao2018sound}  \tabularnewline
Vision                                      &                                  &   Source Separation         & \cite{casanovas2010blind,owens2018audio,gao2018learning,zhao2018sound} \tabularnewline

\bottomrule[2pt]
\end{tabular}
\end{center}
\caption{\label{Table1} Audiovisual learning settings and relevant tasks.}
\end{table}

Audiovisual correspondence is a kind of natural phenomena, which actually comes from the fact that ``Sound is produced by the oscillation of object''.
The simple phenomena provides the possibilities to discover audiovisual appearances and build their complex correlations.
That's why we can match the barking sound to the dog appearance from numerous audio candidates (sound separation) and find the dog appearance according to the barking sound from the complex visual scene (sound source localization).
As usually, the machine model is also expected to possess similarly ability as human.

In the past few years, there have been several works that focus on audiovisual machine learning.
The learning settings and relevant tasks can be categorized into three phases according to source and supervision modality, as shown in Table~\ref{Table1}.
The early works consider that the audio and visual messages of the same entity should have similar class information.
Hence, it is expected to utilize the well-trained model of one modality to supervise the other one without additional annotation.
Such ``teacher-student'' learning fashion has been successfully employed for image classification by sound~\cite{owens2016ambient} and acoustic recognition by vision~\cite{aytar2016soundnet}.

Although the above models have shown promised cross-modal learning capacity, they actually rely on stronger supervision signal than human.
That is, we are not born with a well-trained brain that have recognized kinds of objects or sounds.
Hence, recent (almost concurrent) works propose to train a two-stream network just by given the audiovisual correspondence, as shown in Table~\ref{Table1}.
Arandjelovi{\'c} and Zisserman~\cite{arandjelovic2017look} train their audiovisual model to judge whether the image and audio clip are corresponding.
Although such model is trained without the supervision of any teacher-model, it has learnt highly effective unimodal representation and cross-modal correlation~\cite{arandjelovic2017look}.
Hence, it becomes feasible to execute relevant audiovisual tasks, such as sound localization and source separation.
For the first task, Arandjelovi{\'c} and Zisserman~\cite{arandjelovic2017objects} revise their previous model~\cite{arandjelovic2017look} to find the visual area with the maximum similarity for the current audio clip.
Owens \emph{et al.}~\cite{owens2018audio} propose to adopt the similar model as \cite{arandjelovic2017look} but use 3D convolution network for the visual pathway instead, which can capture the motion information for sound localization.
However, these works rely on simple global correspondence.
When there exist multiple sound-producers in the shown visual modality, it becomes difficult to exactly locate the correct producer.
Recently, Senocak \emph{et al.}~\cite{senocak2018learning} introduce the attention mechanism into the audiovisual model, where the relevant area of visual feature maps learn to attend specific input sound.
However, there still exists another problem that the real-life acoustic environment is usually a mixture of multiple sounds.
To localize the source of specific sound, efficient sound separation is also required.

In the sound separation task, most works propose to reconstruct specific audio streams from manually mixed tracks with the help of visual embedding.
For example, Zhao~\emph{et al.}\cite{zhao2018sound} focus on the musical sound separation, while Casanovas~\emph{et al.}~\cite{casanovas2010blind}, Owens~\emph{et al.}~\cite{owens2018audio}, and Ariel~\emph{et al.}~\cite{ephrat2018looking} perform the separation for mixed speech messages.
However, the real-life sound is more complex and general than the specific imitated examples, which even lacks the groundtruth for the separated sound sources.
Hence, our proposed method jointly disentangles the audio and visual components, and establishes elaborate correspondence between them, which naturally covers both of the sound separation and localization task.


\section{The Proposed Model}

\subsection{Visual and Audio subnet}
\begin{figure}[t]
\centering
\includegraphics[width=8cm]{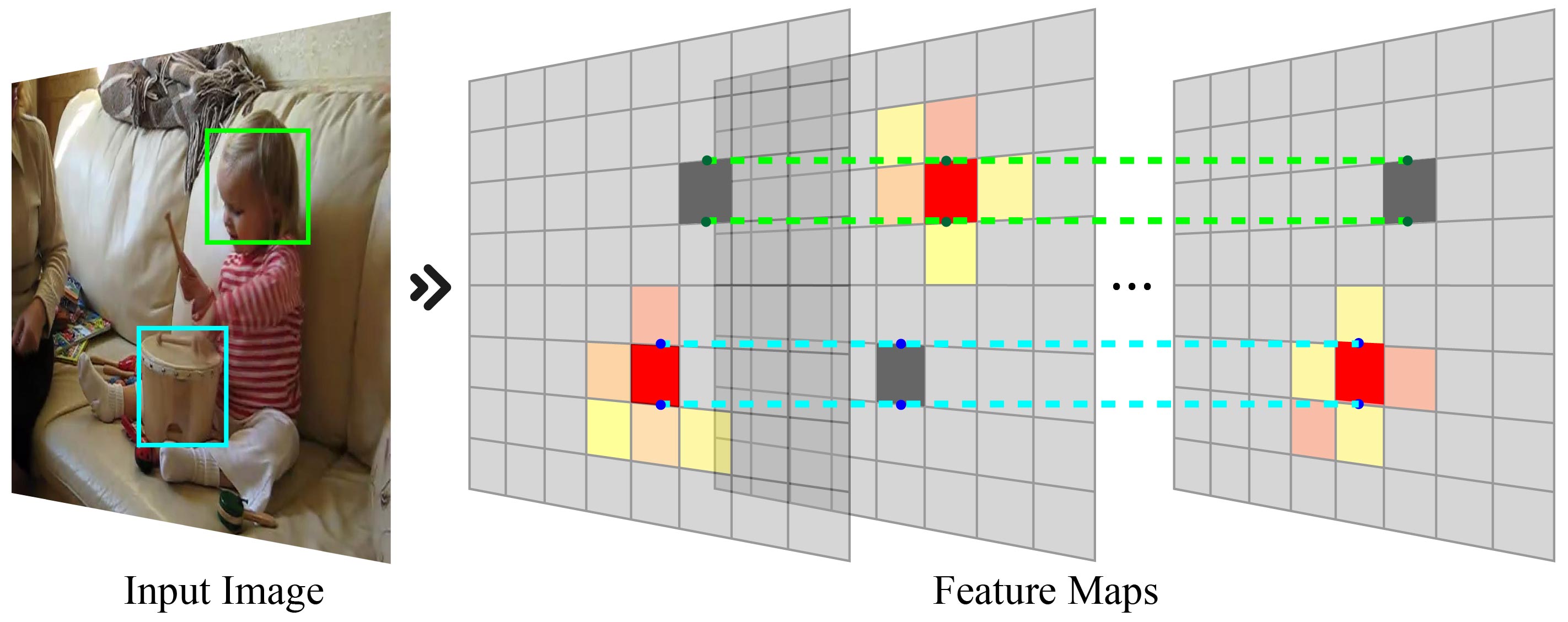}\\
\caption{An illustration of activation distributions. It is obvious that different visual components have distinct activation vectors across the feature maps. Such property helps to distinguish different visual components. Best viewed in color.}\label{featuremap}
\end{figure}

\noindent \textbf{Visual subnet.}
The visual pathway directly adopts the off-the-shelf VGG16 architecture but without the fully connected and softmax layers~\cite{simonyan2014very}.
As the input to the network is resized into $256\times256$ image, the generated 512 feature maps fall into the size of $8\times8$.
To enable the efficient alignment across modalities, the pixel values are scaled into the range of $\left[ { - 1,1} \right]$ that have comparable scale to the log-mel spectrogram of audio signal.
As the associated visual components for the audio signal have been encoded into the feature maps, the corresponding entries across all the maps can be viewed as their feature representations, as shown in Fig.~\ref{featuremap}.
In other words, the original feature maps of $8\times8\times512$ is reshaped into $64\times512$, where each row means the representations for specific visual area.
Hence, the final visual representations become $\left\{ {{u^v_1},{u^v_2},...,{u^v_p}|{u^v_i} \in {R^n}} \right\}$, where $p=64$ and $n=512$.

\noindent \textbf{Audio subnet.}
The audio pathway employs the VGGish model to extract the representations from the input log-mel spectrogram of mono sound~\cite{hershey2017cnn}.
In practice, different from the default configurations in ~\cite{hershey2017cnn}, the input audio clip is extended to 496 frames of 10ms each but other parameters about short-time Fourier transform and mel-mapping are kept.
Hence, the input to the network becomes $496\times64$ log-mel spectrogram, and the corresponding output feature maps become $31 \times 4 \times 512$.
To prepare the audio representation for the second-stage clustering, we also perform the same operation as the visual ones. That is, the audio feature maps are reshaped into $\left\{ {{u^a_1},{u^a_2},...,{u^a_q}|{u^a_i} \in {R^n}} \right\}$, where $q=124$ and $n=512$.

\subsection{Multimodal clustering module}
\begin{figure*}[t]
\centering
\includegraphics[width=17cm]{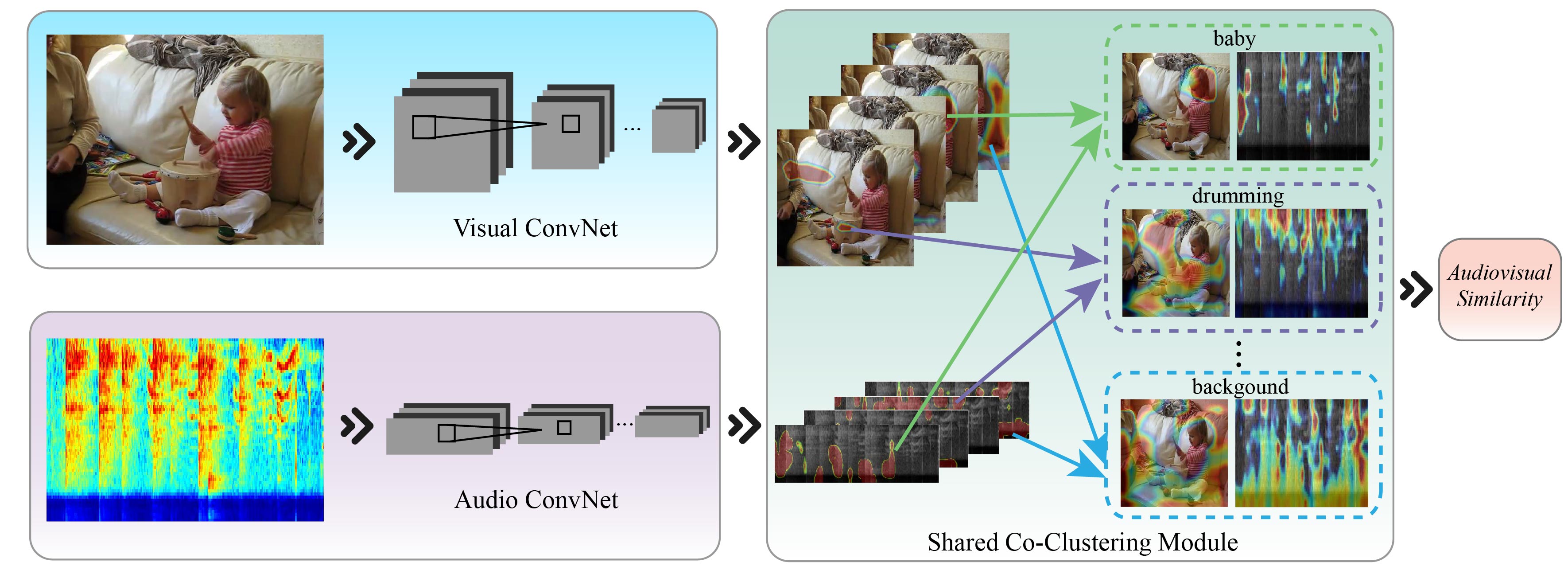}\\
\caption{The diagram of the proposed deep multimodal clustering model. The two modality-specific ConvNets first process the pairwise visual image and audio spectrogram into respective feature maps, then these maps are co-clustered into corresponding components that indicate concrete audiovisual contents, such as baby and its voice, drumming and its sound. Finally, the model takes the similarity across modalities as the supervision for training.}\label{diagram}
\end{figure*}

As the convolutional network shows strong ability in describing the high-level semantics for different modalities~\cite{simonyan2014very,hershey2017cnn,zhang2015character}, we argue that the elements in the feature maps have similar activation probabilities for the same unimodal component, as shown in Fig. \ref{featuremap}. It becomes possible to excavate the audiovisual entities by aggregating their similar feature vectors.
Hence, we propose to cluster the unimodal feature vectors into object-level representations, and align them in the coordinated audiovisual environment, as shown in Fig.~\ref{diagram}.
For simplicity, we take $\left\{ {{u_1},{u_2},...,{u_p}|{u_i} \in {R^n}} \right\}$ for feature representation without regard to the type of modality.

To cluster the unimodal features into $k$ clusters, we propose to perform Kmeans to obtain the centers $C=\left\{ {{c_1},{c_2},...,{c_k}|{c_j} \in {R^m}} \right\}$, where $m$ is the center dimensionality.
Kmeans aims to minimize the within-cluster distance and assign the feature points into $k$-clusters~\cite{jain2010data}, hence, the objective function can be formulated as,
\begin{equation}\label{dcc1}
\mathcal{F}(C) = \sum\limits_{i = 1}^p {\mathop {\min }\limits_{j = 1}^k d\left( {{u_i},{c_j}} \right)},
\end{equation}
where ${\mathop {\min }\limits_{j = 1}^k d\left( {{u_i},{c_j}} \right)}$ means the distance between current point and its closest center.
However, simply introducing Eq.\ref{dcc1} into the deep networks will make it difficult to optimize by gradient descent, as the minimization function in Eq.\ref{dcc1} is a hard assignment of data points for clusters and not differentiable. To solve this intractable problem, one way is to make a soft assignment for each point. Particularly, \emph{Expectation-Maximization} (EM) algorithm for \emph{Gaussian Mixture Models} (GMMs) makes a soft assignment based on the posterior probabilities and converges to a local optimum \cite{kulis2011revisiting}. \par
In this paper, we propose another perspective to transform the hard assignment in Eq.(\ref{dcc1}) to a soft assignment problem and be a differentiable one. The minimization operation in Eq.(\ref{dcc1}) is approximated via utilizing the following equation,
\begin{equation}\label{dcc2}
\max \left\{ {{d_{i1}},{d_{i2}},...,{d_{ik}}} \right\} \approx \frac{1}{z}\log \left( {\sum\limits_{j = 1}^k {{e^{{d_{ij}}z}}} } \right),
\end{equation}
where $z$ is a parameter about magnitude and $d_{ij}=d\left( {{u_i},{c_j}} \right)$ for simplicity. Eq.\ref{dcc2} shows that the maximum value of a given sequence can be approximated by the $log$-summation of corresponding exponential functions.
Intuitively, the differences in the original sequence are amplified sharply with the exponential projection, which tends to ignore the tiny ones and remain the largest one.
Then, the reversed logarithm projection gives the approximated maximum value. The rigorous proof for Eq.\ref{dcc2} can be found in the materials.

As we aim to find the minimum value of the distance sequence, Eq. \ref{dcc2} is modified into
\begin{equation}\label{dcc4}
\min \left\{ {{d_{i1}},{d_{i2}},...,{d_{ik}}} \right\} \approx  - \frac{1}{z}\log \left( {\sum\limits_{j = 1}^k {{e^{ - {d_{ij}}z}}} } \right).
\end{equation}
Then, the objective function of clustering becomes
\begin{equation}\label{dcco}
\mathcal{F}\left( C \right) =  - \frac{1}{z}\sum\limits_{i = 1}^p {\log \left( {\sum\limits_{j = 1}^k {{e^{ - {d_{ij}}z}}} } \right)}.
\end{equation}
As Eq.~\ref{dcco} is differentiable everywhere, we can directly compute the derivative w.r.t. each cluster center.
Concretely, for the center $c_j$, the derivative is written as
\begin{equation}\label{dcc5}
\frac{{\partial \mathcal{F} }}{{\partial {c_j}}} = \sum\limits_{i = 1}^p {\frac{{{e^{ - {d_{ij}}z}}}}{{\sum\limits_{l = 1}^k {{e^{ - {d_{il}}z}}} }}\frac{{\partial {d_{ij}}}}{{\partial {c_j}}}}  = \sum\limits_{i = 1}^p {{s_{ij}}\frac{{\partial {d_{ij}}}}{{\partial {c_j}}}},
\end{equation}
where ${s_{ij}} = \frac{{{e^{ - {d_{ij}}z}}}}{{\sum\nolimits_{l = 1}^k {{e^{ - {d_{il}}z}}} }} = softmax\left( { - {d_{ij}}z} \right)$.
The softmax coefficient performs like the soft-segmentation over the whole visual area or audio spectrogram for different centers, and we will give more explanation about it in the following sections.

In practice, the distance $d_{ij}$ between each pair of feature point $u_i$ and center $c_j$ can be achieved in different ways, such as Euclidean distance, cosine proximity, etc.
In this paper, inspired by the capsule net\footnote{A discussion about capsule and DMC is provided in the materials.}~\cite{sabour2017dynamic, wang2018optimization}, we choose the inner-product for measuring the agreement, i.e., ${d_{ij}} =  - \left\langle {{u_i},\frac{{{c_j}}}{{\left\| {{c_j}} \right\|}}} \right\rangle $. By taking it into Eq.~\ref{dcc5} and setting the derivative to zero, we can obtain\footnote{Detailed derivation is shown in the materials.}
\begin{equation}\label{dcc6}
\frac{{c_j}}{{\left\| {c_j} \right\|}} = \frac{{\sum\limits_{i = 1}^p {s_{ij}{u_i}} }}{{\left\| {\sum\limits_{i = 1}^p {s_{ij}{u_i}} } \right\|}},
\end{equation}
which means the center and the integrated features lie in the same direction.
As the coefficients $s_{\cdot j}$ are the softmax values of distances, the corresponding center $c_j$ emerges in the comparable scope as the features and Eq.~\ref{dcc6} is approximatively computed as $c_j = \sum\limits_{i = 1}^p {s_{ij}{u_i}}$ for simplicity.
However, there remains another problem that the computation of $s_{ij}$ depends on the current center $c_j$, which makes it difficult to get the direct update rules for the centers.
Instead, we choose to alternatively update the coefficient ${s_{ij}^{\left( r \right)}}$ and center $c_j^{\left( {r + 1} \right)}$, i.e.,
\begin{equation}\label{dcc7}
c_j^{\left( {r + 1} \right)} = \sum\limits_{i = 1}^p {s_{ij}^{\left( r \right)}{u_i}}.
\end{equation}
Actually, the updating rule is much similar to the EM algorithm that maximizes posterior probabilities in GMMs \cite{bishop2006pattern}. Specifically, the first step is the \emph{expectation} step or E step, which uses the current parameters to evaluate the posterior probabilities, i.e.,  re-assigns data points to the centers. The second step is the \emph{maximization} step or M step that aims to re-estimate the means, covariance, and mixing coefficients, i.e., update the centers in Eq.(\ref{dcc7}).

The aforementioned clusters indicate a kind of soft assignment (segmentation) over the input image or spectrogram, where each cluster mostly corresponds to certain content (e.g., baby face and drum in image, voice and drumbeat in sound in Fig.~\ref{diagram}), hence they can be viewed as the distributed representations of each modality.
And we argue that audio and visual messages should have similar distributed representations when they jointly describe the same natural scene.
Hence, we propose to perform different center-specific projections $\left\{W_1, W_2,...,W_k\right\}$ over the audio and visual messages to distinguish the representations of different audiovisual entities, then cluster these projected features into the multimodal centers for seeking concrete audiovisual contents.
Formally, the distance $d_{ij}$ and center updating become ${d_{ij}} =  - \left\langle {{W_j}{u_i},\frac{{{c_j}}}{{\left\| {{c_j}} \right\|}}} \right\rangle $ and $c_j^{\left( {r + 1} \right)} = \sum\limits_{i = 1}^p {s_{ij}^{\left( r \right)}{W_j}{u_i}}$, where the projection matrix $W_j$ is shared across modalities and considered as the association with concrete audiovisual entity. Moreover, $W_j$ also performs as the magnitude parameter $z$ when computing the distance $d_{ij}$. We show the complete multimodal clustering in Algorithm 1.

We employ the cosine proximity to measure the difference between audiovisual centers, i.e., $s\left( {c_i^a,c_i^v} \right)$, where $c_i^a$ and $c_i^v$ are the $i$-center for audio and visual modality, respectively.
To efficiently train the two-stream audiovisual network, we employ the max-margin loss to encourage the network to give more confidence to the realistic image-sound pair than mismatched ones,
\begin{equation}\label{dcc8}
loss = \sum\limits_{i = 1,i \ne j}^k {\max \left( {0,s\left( {c_j^a,c_i^v} \right) - s\left( {c_i^a,c_i^v} \right) + \Delta } \right)},
\end{equation}
where $\Delta$ is a margin hyper-parameter and $c_j^a$ means the negative audio sample for the positive audiovisual pair of $\left( {c_i^a,c_i^v} \right)$.
In practice, the negative example is randomly sampled from the training set but different from the positive one.
The Adam optimizer with the learning rate of ${10^{ - 4}}$ is used. Batch-size of 64 is selected for optimization. And we train the audiovisual net for 25,000 iterations, which took 3 weeks on one K80 GPU card.

\begin{algorithm}[h]
\caption{Deep Multimodal Clustering Algorithm}
    \begin{algorithmic}[1]
\REQUIRE
The feature vectors for each modality: \\
 $\left\{ {u_1^a,u_2^a,...,u_q^a|u_i^a \in {R^n}} \right\}$, $\left\{ {u_1^v,u_2^v,...,u_p^v|u_i^v \in {R^n}} \right\}$
\ENSURE
The center vectors for each modality: \\
 $\left\{ {c_1^a,c_2^a,...,c_k^a|c_j^a \in {R^m}} \right\}$, $\left\{ {c_1^v,c_2^v,...,c_k^v|c_j^v \in {R^m}} \right\}$

 Initialize the distance $d_{ij}^a = d_{ij}^v = 0$
    \FOR{$t=1$ to $T$, $x$ in $\left\{ {a,v} \right\}$}
        \FOR{$i=1$ to $q\left(p\right)$, $j=1$ to $k$}
         \STATE Update weights: $s_{ij}^x = softmax\left( { - d_{ij}^x} \right)$
         \STATE Update centers: $ c_j^x = \sum\limits_{i = 1}^p {s_{ij}^x{W_j}u_i^x}$
         \STATE Update distances: $d_{ij}^x =  - \left\langle {{W_j}u_i^x,\frac{{c_j^x}}{{\left\| {c_j^x} \right\|}}} \right\rangle$
         \ENDFOR
    \ENDFOR

    \label{code:recentEnd}
    \end{algorithmic}
\end{algorithm}




\section{Feature Evaluation}
Ideally, the unimodal networks should have learnt to respond to different objects or sound scenes after training the DMC model.
Hence, we propose to evaluate the learned audio and visual representations of the CNN internal layers.
For efficiency, the DMC model is trained with 400K unlabeled videos that are randomly sampled from the SoundNet-Flickr dataset~\cite{aytar2016soundnet}.
The input audio and visual message are the same as~\cite{aytar2016soundnet}, where pairs of 5s sound clip and corresponding image are extracted from each video with no overlap.
Note that, the constituted $\sim$1.6M audiovisual pairs are about 17 times less than the ones in $L^3$~\cite{arandjelovic2017look} and 5 times less than SoundNet~\cite{aytar2016soundnet}.

\subsection{Audio Features}
The audio representation is evaluated in the complex environmental sound classification task. The adopted ESC-50 dataset~\cite{piczak2015esc} is a collection of 2000 audio clips of 5s each. They are equally partitioned into 50 categories. Hence, each category contains 40 samples. For fairness, each sample is also partitioned into 1s audio excerpts for data argumentation~\cite{aytar2016soundnet}, and these overlapped subclips constitute the audio inputs to the VGGish network. The mean accuracy is computed over the five leave-one-fold-out evaluations. Note that, the human performance on this dataset is 0.813.

The audio representations are extracted by pooling the feature maps\footnote{Similarly with SoundNet~\cite{aytar2016soundnet}, we evaluate the performance of different VGGish layers and select conv4\_1 as the extraction layer.}. And a multi-class one-vs-all linear SVM is trained with the extracted audio representations. The final accuracy of each clip is the mean value of its subclip scores.
To be fair, we also modify the DMC model into the ``teacher-student'' scheme ($\ddag$DMC) where the VGG net is pretrained with ImageNet and kept fixed during training.
In Table \ref{Table2} (a), it is obvious that the DMC model exceeds all the previous methods except ~\emph{Audio-Visual Temporal Synchronization} (AVTS)~\cite{korbar2018co}.
Such performance is achieved with less training data (just 400K videos), which confirms that our model can utilize more audiovisual correspondences in the unconstrained videos to effectively train the unimodal network.
We also note that AVTS is trained with the whole 2M+ videos in~\cite{aytar2016soundnet}, which is 5 times more than DMC.
Even so, DMCs still outperform AVTS on the DCASE2014 benchmark dataset (More details can be found in the materials).
And the cross-modal supervision version $\ddag$DMC improves the accuracy further, where the most noticeable point is that $\ddag$DMC outperforms human~\cite{piczak2015esc} (82.6$\%$ vs 81.3$\%$).
Hence, it verifies that the elaborative alignment efficiently works and the audiovisual correspondence indeed helps to learn the unimodal perception.

\begin{table}[t]
\begin{center}
\begin{minipage}[h]{0.5\linewidth}
\centerline{(a) ESC-50}
  \newcolumntype{C}[1]{>{\centering}p{#1}}
\begin{tabular}{C{2.5cm}C{1.1cm}}
\toprule[2pt]
  Methods  &  Accuracy
\tabularnewline\midrule[1pt]
Autoencoder~\cite{aytar2016soundnet} & 0.399 \tabularnewline
Rand. Forest~\cite{piczak2015esc}    & 0.443 \tabularnewline
  ConvNet~\cite{piczak2015environmental}    & 0.645 \tabularnewline

  \hline

  SoundNet~\cite{aytar2016soundnet} & 0.742 \tabularnewline

    $L^3$~\cite{arandjelovic2017look}  &   0.761 \tabularnewline

  $\dag L^3$~\cite{arandjelovic2017look}    &  0.793 \tabularnewline

  $\dag$AVTS~\cite{korbar2018co}           & \underline{0.823} \tabularnewline

  DMC    & 0.798 \tabularnewline

  $\ddag$DMC    &  \textbf{0.826} \tabularnewline

  \hline
 \emph{Human  Perfor.} & 0.813 \tabularnewline

\bottomrule[2pt]
\end{tabular}
\end{minipage}
\hfill
\begin{minipage}[h]{0.4\linewidth}
\centerline{(b) Pascal VOC 2007}
  \newcolumntype{C}[1]{>{\centering}p{#1}}
\begin{tabular}{C{1.8cm}C{1.1cm}}
\toprule[2pt]
  Methods  &  Accuracy
\tabularnewline\midrule[1pt]
  Taxton.~\cite{owens2016ambient} &  0.375   \tabularnewline
  Kmeans~\cite{krahenbuhl2015data}     &   0.348   \tabularnewline
  Tracking~\cite{wang2015unsupervised}       & 0.422  \tabularnewline
  Patch.~\cite{doersch2015unsupervised}    & 0.467 \tabularnewline
  Egomotion~\cite{agrawal2015learning}    & 0.311 \tabularnewline
  \hline
  Sound(spe.)~\cite{owens2016ambient} & 0.440 \tabularnewline
  Sound(clu.)~\cite{owens2016ambient} & 0.458 \tabularnewline
  Sound(bia.)~\cite{owens2016ambient} & 0.467 \tabularnewline
    DMC  &   \textbf{0.514} \tabularnewline
\hline
  ImageNet    &  \underline{0.672} \tabularnewline

\bottomrule[2pt]
\end{tabular}
\end{minipage}
\end{center}
\caption{\label{Table2}Acoustic Scene Classification on ESC-50~\cite{piczak2015esc} and Image Classification on Pascal VOC 2007\cite{everingham2010pascal}. (a) For fairness, we provide a weakened version of $L^3$ that is trained with the same audiovisual set as ours, while $\dag L^3$ is trained with more data in~\cite{arandjelovic2017look}. $\dag$AVTS is trained with the whole SoundNet-Flickr dataset~\cite{aytar2016soundnet}. $\ddag$DMC takes supervision from the well-trained vision network for training the audio subnet. (b) The shown results are the best ones reported in~\cite{owens2016ambient} except the ones with FC features.}
\end{table}

\subsection{Visual Features}
The visual representation is evaluated in the object recognition task. The chosen PASCAL VOC 2007 dataset contains 20 object categories that are collected in realistic scenes~\cite{everingham2010pascal}.
We perform global-pooling over the conv5\_1 features of VGG16 net to obtain the visual features. A multi-class one-vs-all linear SVM is also employed as the classifier, and the results are evaluated using \emph{Mean Average Precision} (mAP).
As the DMC model does not contain the standard FC layer as previous works, the best conv/pooling features of other methods are chosen for comparison, which have been reported in~\cite{owens2016ambient}.
To validate the effectiveness of multimodal clustering in DMC, we choose to compare with the visual model in~\cite{owens2016ambient}, which treats the separated sound clusters as object indicators for visual supervision.
In contrast, DMC model jointly learns the audio and visual representation rather than the above single-flow from sound to vision, hence it is more flexible for learning the audiovisual correspondence.
As shown in Table~\ref{Table2} (b), our model indeed shows noticeable improvement over the simple cluster supervision, even its multi-label variation (in binary)~\cite{owens2016ambient}.
Moreover, we also compare with the pretrained VGG16 net on ImageNet.
But, what surprises us is that the DMC model is comparable the human performance in acoustic classification but it has a large gap with the image classification benchmark.
Such differences may come from the complexity of visual scene compared with the acoustic ones.
Nevertheless, our model still provides meaningful insights in learning effective visual representation via audiovisual correspondence.

\begin{figure}[t]
\centering
\includegraphics[width=8cm]{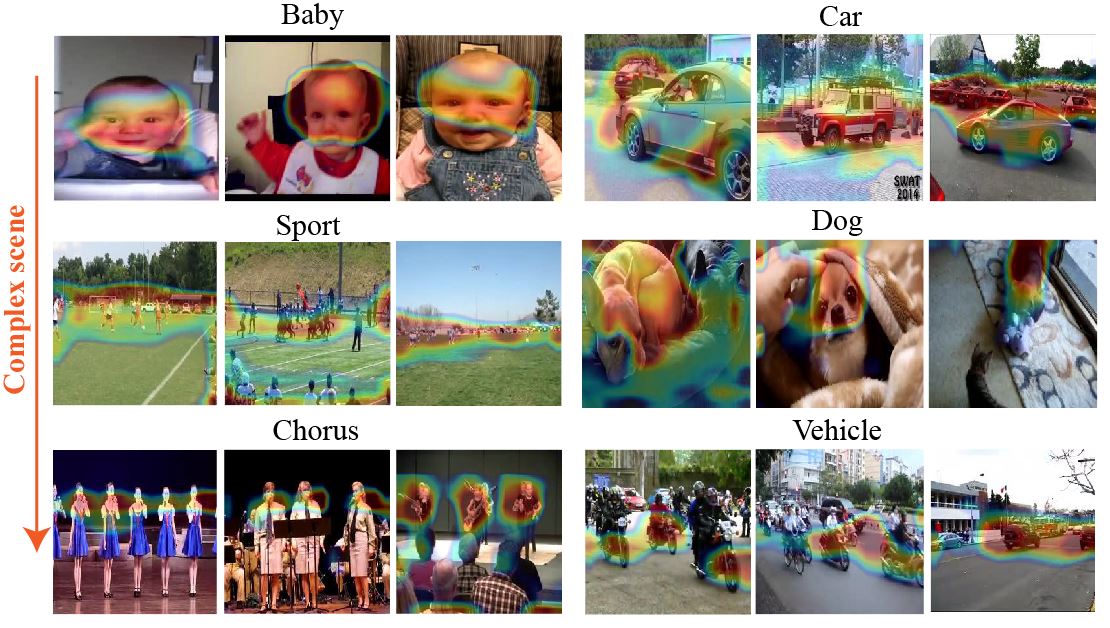}\\
\caption{Qualitative examples of sound source localization. After feeding the audio and visual messages into the DMC model, we visualize the soft assignment that belongs to the most related visual cluster to the audio messages. Note that, the visual scene becomes more complex from top to bottom, and the label is just for visualization purpose.}\label{source}
\end{figure}

\section{Audiovisual Evaluation}

\subsection{Single Sound Localization}
In this task, we aim to localize the sound source in the visual scene as~\cite{arandjelovic2017objects,owens2018audio}, where the simple case of single source is considered.
As only one sound appears in the audio track, the generated audio features should share identical center.
In practice, we perform average-pooling over the audio centers into $c^a$, then compare it with all the visual centers via cosine proximity, where the number of visual centers is set to 2 (i.e., sound-maker and other components) in this simple case.
And the visual center with the highest score is considered as the indicator of corresponding sound source.
In order to visualize the sound source further, we resort to the soft assignment of the selected visual center $c^v_j$, i.e., $s^v_{\cdot j}$.
As the assignment $s^v_{ij} \in \left[ {0,1} \right]$, the coefficient vector $s^v_{\cdot j}$ is reshaped back to the size of original feature map and viewed as the heatmap that indicates the cluster property.

In Fig.~\ref{source}, we show the qualitative examples with respect to the sound-source locations of different videos from the SoundNet-Flickr dataset.
It is obvious that the DMC model has learnt to distinguish different visual appearances and correlate the sound with corresponding visual source, although the training phase is entirely performed in the unsupervised fashion.
Concretely, in the simple scene, the visual source of baby voice and car noise are easy to localize.
When the visual scene becomes more complex, the DMC model can also successfully localize the corresponding source.
The dog appearance highly responses to the barking sound while the cat does not.
In contrast to the audience and background, only the onstage choruses response to the singing sound.
And the moving vehicles are successfully localized regardless of the driver or other visual contents in the complex traffic environment.

Apart from the qualitative analysis, we also provide the quantitative evaluation. We directly adopt the annotated sound-sources dataset~\cite{senocak2018learning}, which are originally collected from the SoundNet-Flickr dataset. This sub-dataset contains 2,786 audio-image pairs, where the sound-maker of each pair is individually located by three subjects. 250 pairs are randomly sampled to construct the testing set (with single sound).
By setting an arbitrary threshold over the assignment $s^v_{\cdot j}$, we can obtain a binary segmentation over the visual objects that probably indicate the sound locations.
Hence, to compare the automatic segmentation with human annotations, we employ the ~\emph{consensus Intersection over Union} (cIoU) and corresponding AUC area in~\cite{senocak2018learning} as the evaluation metric.
As shown in Table.~\ref{Table5}, the proposed DMC model is compared with the recent sound location model with attention mechanism~\cite{senocak2018learning}.
First, it is obvious that DMC shows superior performance over the unsupervised attention model. Particularly, when the cIoU threshold becomes larger (i.e., 0.7), DMC even outperforms the supervised ones.
Second, apart from the most related visual centers to the sound track, the unrelated one is also evaluated. The large decline of the unrelated center indicates that the clustering mechanism in DMC can effectively distinguish different modality components and exactly correlate them among different modalities.


\subsection{Real-Life Sound Event Detection}
In this section, in contrast to specific sound separation task\footnote{The sound separation task mostly focuses specific task scenarios and need effective supervision from the original sources\cite{casanovas2010blind,owens2018audio,zhao2018sound}, which goes beyond our conditions.}, we focus on a more general and complicated sound task, i.e., multisource SED.
In the realistic environment, multiple sound tracks usually exist at the same time, i.e.,  the street environment may be a mixture of people speaking, car noise, walking sound, brakes squeaking, etc.
It is expected to detect the existing sounds in every moment, which is much more challenging than the previous single acoustic recognition~\cite{Heittola2017}.
Hence, it becomes more valuable to evaluate the ability of DMC in learning the effective representation of multi-track sound.
In the DCASE2017 acoustic challenges, the third task\footnote{ http://www.cs.tut.fi/sgn/arg/dcase2017/challenge/task-sound-event-detection-in-real-life-audio} is exactly the multisource SED.
The audio dataset used in this task focuses on the complex street acoustic scenes that consist of different traffic levels and activities.
The whole dataset is divided for development and evaluation, and each audio is 3-5 minutes long.
Segment-based F-score and error rate are calculated as the evaluation metric.
\begin{table}[t]
\begin{center}
\begin{tabular}{cccc}
\toprule[2pt]
  Methods  &  cIoU(0.5) & cIoU(0.7) & AUC
\tabularnewline\midrule[1pt]

    Random                                                                   &   12.0      &   -    &    32.3    \tabularnewline
    Unsupervised$\dag$~\cite{senocak2018learning}       &   52.4      &  -     &  51.2   \tabularnewline
    Unsupervised~\cite{senocak2018learning}       &    66.0      &  $\sim$18.8     &  55.8    \tabularnewline
    Supervised~\cite{senocak2018learning}           &    80.4      &  $\sim$25.5     &    60.3    \tabularnewline
    Sup.+Unsup.~\cite{senocak2018learning}        &    82.8      &  $\sim$28.8    &   62.0    \tabularnewline
  \hline
  DMC (unrelated)                                                    &  10.4 & 5.2 & 21.1  \tabularnewline
  DMC (related)                                                        &  67.1 & 26.2  & 56.8  \tabularnewline

\bottomrule[2pt]
\end{tabular}
\end{center}
\caption{\label{Table5} The evaluation of sound source localization. The cIoUs with threshold 0.5 and 0.7 are shown. The area under the cIoU curve by varying the threshold from 1 to 0 (AUC) is also provided.
The unsupervised$\dag$ method in \cite{senocak2018learning} employs a modified attention mechanism.
}
\end{table}

As our model provides elaborative visual supervision for training the audio subnet, the corresponding audio representation should provide sufficient description for the multitrack sound.
To validate the assumption, we directly replace the input spectrum with our generated audio representation in the baseline model of MLP~\cite{Heittola2017}.
As shown in Table.~\ref{Table4}, the DMC model is compared with the top five methods in the challenge, the audiovisual net $L^3$~\cite{arandjelovic2017look}, and the VGGish net~\cite{hershey2017cnn}.
It is obvious that our model takes the first place on F1 metric and is comparable to the best model in error rate.
Specifically, there are three points we should pay attention to.
First, by utilizing the audio representation of DMC model instead of raw spectrum, we can have a noticeable improvement.
Such improvement indicates that the correspondence learning across modalities indeed provides effective supervision in distinguishing different audio contents.
Second, as the $L^3$ net simply performs global matching between audio and visual scene without exploring the concrete content inside, it fails to provide effective audio representation for multisource SED.
Third, although the VGGish net is trained on a preliminary version of YouTube-8M (with labels) that is much larger than our training data, our model still outperforms it.
This comes from the more efficient audiovisual correspondence learning of DMC model.

\begin{table}[t]
\begin{center}
\begin{tabular}{ccc}
\toprule[2pt]
  Methods  &  Segment F1 & Segment Error
\tabularnewline\midrule[1pt]

  J-NEAT-E~\cite{Kroos2017}                           &  44.9  & 0.90  \tabularnewline
  SLFFN~\cite{Kroos2017}                                &  43.8  & 1.01  \tabularnewline
  ASH~\cite{Adavanne2017}                                    &  41.7   & \textbf{0.79} \tabularnewline
  MICNN~\cite{Jeong2017}                              &  40.8   & 0.81 \tabularnewline
    MLP~\cite{Heittola2017}                  &   42.8 & 0.94   \tabularnewline
  $\S$MLP~\cite{Heittola2017}       &  39.1 & 0.90          \tabularnewline

  \hline
  $L^3$~\cite{arandjelovic2017look}                         &   43.24     & 0.89  \tabularnewline
  VGGish~\cite{hershey2017cnn}                              &  50.96   & 0.86 \tabularnewline
  DMC                                              &  \textbf{52.14} & 0.83 \tabularnewline

\bottomrule[2pt]
\end{tabular}
\end{center}
\caption{\label{Table4}Real life sound event detection on the evaluation dataset of DCASE 2017 Challenge. We choose the default STFT parameters of 25ms window size and 10 window hop~\cite{hershey2017cnn}. The same parameters are also adopted by $\S$MLP, $L^3$, and VGGish, while other methods adopt the default parameters in~\cite{Heittola2017}.}
\end{table}

\subsection{Audiovisual Understanding}
\begin{figure}[t]
\centering
\includegraphics[width=8cm]{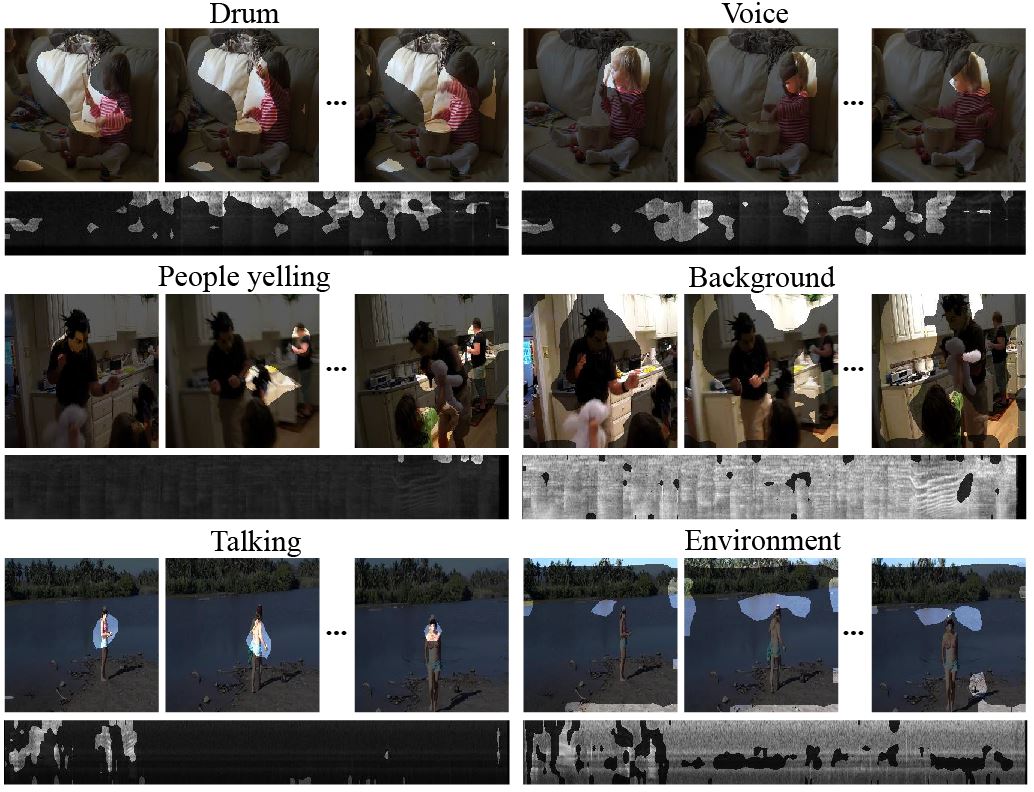}\\
\caption{Qualitative examples of complex audiovisual understanding. We first feed the audiovisual messages into the DMC model, then the corresponding audiovisual clusters are captured and shown, where the assignments are binarized into the masks over each modality via a threshold of 0.7. The labels in the figure indicate the learned audiovisual content, which is not used in the training procedure.}\label{avu}
\end{figure}

As introduced in the Section 1, the real-life audiovisual environment is unconstrained, where each modality consists of multiple instances or components, such as speaking, brakes squeaking, walking sound in the audio modality and building, people, cars, road in the visual modality of the street environment.
Hence, it is difficult to disentangle them within each modality and establish exact correlations between modalities, i.e., audiovisual understanding.
In this section, we attempt to employ the DMC model to perform the audiovisual understanding in such cases where only the qualitative evaluation is provided due to the absent annotations.
To illustrate the results better, we turn the soft assignment of clustering into a binary map via a threshold of 0.7.
And Fig.~\ref{avu} shows the matched audio and visual clustering results of different real-life videos, where the sound is represented in spectrogram.
In the ``baby drums'' video, the drumming sound and corresponding motion are captured and correlated, meanwhile the baby face and people voice are also picked out from the intricate audiovisual content.
These two distinct centers jointly describe the audiovisual structures.
In more complex indoor and outdoor environments, the DMC model can also capture the people yelling and talking sound from background music and loud environment noise by clustering the audio feature vectors, and correlate them with the corresponding sound-makers (i.e., the visual centers) via the shared projection matrix.
However, there still exist some failure cases.
Concretely, the out-of-view sound-maker is inaccessible for current visual clustering.
In contrast, the DMC model improperly correlates the background music with kitchenware in the second video.
Similarly, the talking sound comes from the visible woman and out-of-view photographer in the third video, but our model simply extracts all the human voice and assigns them to the visual center of woman.
Such failure cases also remind us that the real-life audiovisual understanding is far more difficult than what we have imagined.
Moreover, to perceive the audio centers more naturally, we reconstruct the audio signal from the masked spectrogram information and show them in the released video demo.

\section{Discussion}
In this paper, we aim to explore the elaborate correspondence between audio and visual messages in unconstrained environment by resorting to the proposed deep multimodal clustering method.
In contrast to the previous rough correspondence, our model can efficiently learn more effective audio and visual features, which even exceed the human performance.
Further, such elaborate learning contributes to noticeable improvements in the complicated audiovisual tasks, such as sound localization, multisource SED, and audiovisual understanding.

Although the proposed DMC shows considerable superiority over other methods in these tasks, there still remains one problem that the number of clusters $k$ is pre-fixed instead of automatically determined.
When there is single sound, it is easy to set $k=2$ for foreground and background. But when multiple sound-makers emerge, it becomes difficult to pre-determine the value of $k$.
Although we can obtain distinct clusters after setting $k=10$ in the audiovisual understanding task, more reliable method for determining the number of audiovisual components is still expected~\cite{ray1999determination}, which will be focused in the future work.



\section{Acknowledgement}
This work was supported in part by the National Natural Science Foundation of China grant under number 61772427 and 61751202. We thank Jianlin Su for the constructive opinion, and thank Zheng Wang and reviewers for refreshing the paper.

{\small
\bibliographystyle{ieee}
\bibliography{dccl}
}

\clearpage

\section{Appendix}

\subsection{Approximated Maximization Function}
Although the maximization function is not differentiable, it can be approximated via the following equation,
\begin{equation}\label{proof1}
\max \left\{ {{d_{i1}},{d_{i2}},...,{d_{ik}}} \right\} \approx \mathop {\lim }\limits_{z \to  + \infty } \frac{1}{z}\log \left( {\sum\limits_{j = 1}^k {{e^{{d_{ij}}z}}} } \right),
\end{equation}
where $z$ is a hype-parameter that controls the precision of approximation.
Instead of the above multi-variable case, we first consider the maximization function of two variables $\left\{ {{x_1},{x_2}} \right\}$.
Actually, it is well known that
\begin{equation}\label{proof2}
\begin{array}{l}
\max \left\{ {{x_1},{x_2}} \right\} = \frac{1}{2}\left( {\left| {{x_1} + {x_2}} \right| + \left| {{x_1} - {x_2}} \right|} \right),\\
{\rm{~~~~~~~~~~~~~}}s.t. {\rm{~~~}}{x_1} \ge 0,{x_2} \ge 0, \\
\end{array}
\end{equation}
Hence the approximation for maximization is turned for the absolute value function $f\left( x \right) = \left| x \right|$. As the derivative function of $f\left( x \right)$ is
$f'\left( x \right) = \left\{ {\begin{array}{*{20}{c}}
{ + 1,{\rm{~~}} x \ge 0}\\
{ - 1,{\rm{~~}}x < 0}
\end{array}}, \right.$
it can be directly replaced by the adaptive tanh function~\cite{li2017deep}, i.e., $f'\left( x \right) = \mathop {\lim }\limits_{z \to  + \infty } \frac{{{e^{zx}} - {e^{ - zx}}}}{{{e^{zx}} + {e^{ - zx}}}}$.
Then, we can obtain the approximated absolute value function via integral
\begin{equation}\label{proof3}
f\left( x \right) =  \mathop {\lim }\limits_{z \to  + \infty } \frac{1}{z}\log \left( {{e^{zx}} + {e^{ - zx}}} \right).
\end{equation}
Hence, the maximization function over two variables can be written as
\begin{equation}\label{proof4}
\begin{array}{l}
\max \left\{ {{x_1},{x_2}} \right\}\\
= \mathop {\lim }\limits_{z \to  + \infty } \frac{1}{{2z}}\log \left( {{e^{2z{x_1}}} + {e^{2z{x_2}}} + {e^{ - 2z{x_1}}} + {e^{ - 2z{x_2}}}} \right).\\
\end{array}
\end{equation}

As $z \to  + \infty$ and ${x_1} \ge 0,{x_2} \ge 0$, Eq.~\ref{proof4} can be approximated into
\begin{equation}\label{proof5}
\max \left\{ {{x_1},{x_2}} \right\} \approx \mathop {\lim }\limits_{z \to  + \infty } \frac{1}{z}\log \left( {{e^{z{x_1}}} + {e^{z{x_2}}}} \right).
\end{equation}
At this point, the maximization function has become differentiable for two variables. And it can also be extended to three more variables.
Concretely, for three variables $\left\{ {{x_1},{x_2},{x_3}} \right\}$, let $c = \max \left\{ {{x_1},{x_2}} \right\}$, then
\begin{equation}\label{proof7}
\begin{array}{l}
\max \left\{ {{x_1},{x_2},{x_3}} \right\} = \max \left\{ {{c},{x_3}} \right\} \\
{\rm{~~~~~~~~~~~~~~~~~~~~~~~~~~~~ }}\approx \mathop {\lim }\limits_{z \to  + \infty } \frac{1}{z}\log \left( {{e^{\log \left( {{e^{z{x_1}}} + {e^{z{x_2}}}} \right)}} + {e^{z{x_3}}}} \right) \\
{\rm{~~~~~~~~~~~~~~~~~~~~~~~~~~~~ = }}\mathop {\lim }\limits_{z \to  + \infty } \frac{1}{z}\log \left( {{e^{z{x_1}}} + {e^{z{x_2}}} + {e^{z{x_3}}}} \right).
\end{array}
\end{equation}
Hence, for multivariable, we can have
\begin{equation}\label{proof8}
\max \left\{ {{x_1},{x_2},...,{x_n}} \right\} \approx \mathop {\lim }\limits_{z \to  + \infty } \frac{1}{z}\log \left( {\sum\limits_{i = 1}^n {{e^{z{x_i}}}} } \right).
\end{equation}

\subsection{Derivation of Eq.~\ref{dcc6}}
To substitute ${d_{ij}} =  - \left\langle {{u_i},\frac{{{c_j}}}{{\left\| {{c_j}} \right\|}}} \right\rangle $ into $\sum\limits_{i = 1}^n {s_{ij}\frac{{\partial d_{ij}}}{{\partial c_j}}}  = 0$,
we first give the derivative of ${d_{ij}}$ w.r.t. $c_j$,
\begin{equation}\label{proof21}
\frac{{\partial {d_{ij}}}}{{\partial {c_j}}} =  - \frac{{\partial \left( {\frac{{u_i^T{c_j}}}{{\left\| {{c_j}} \right\|}}} \right)}}{{\partial {c_j}}} =  - \frac{{{u_i}}}{{\left\| {{c_j}} \right\|}} + u_i^T{c_j} \cdot \frac{{{c_j}}}{{{{\left\| {{c_j}} \right\|}^3}}}.
\end{equation}
Then, by taking Eq.~\ref{proof21} into $\sum\limits_{i = 1}^n {s_{ij}\frac{{\partial d_{ij}}}{{\partial c_j}}}  = 0$, we can have
\begin{equation}\label{proof22}
\sum\limits_{i = 1}^n {s_{ij}\frac{{u_i^Tc_j}}{{\left\| {c_j} \right\|}}}  \cdot \frac{{c_j}}{{\left\| {c_j} \right\|}} = \sum\limits_{i = 1}^n {s_{ij}{u_i}}.
\end{equation}
By taking the modulus of expression in both sides of Eq.~\ref{proof22}, we can have
\begin{equation}\label{proof24}
\left\| {\sum\limits_{i = 1}^n {s_{ij}\frac{{u_i^Tc_j}}{{\left\| {c_j} \right\|}}} } \right\| \cdot \left\| {\frac{{c_j}}{{\left\| {c_j} \right\|}}} \right\| = \left\| {\sum\limits_{i = 1}^n {s_{ij}{u_i}} } \right\|.
\end{equation}
As $ \left\| {\frac{{c_j}}{{\left\| {c_j} \right\|}}} \right\|{\rm{ = }}1$, Eq. \ref{proof24} becomes
\begin{equation}\label{proof25}
\left\| {\sum\limits_{i = 1}^n {s_{ij}{u_i}} } \right\|{\rm{ = }}\left\| {\frac{{\sum\limits_{i = 1}^n {s_{ij}u_i^T}  \cdot c_j}}{\left\| {c_j} \right\|}} \right\|{\rm{ = }}\left\| {\sum\limits_{i = 1}^n {s_{ij}{u_i}} } \right\|\left| {\cos \theta } \right|
\end{equation}

As ${d_{ij}} =  - \left\langle {{u_i},\frac{{{c_j}}}{{\left\| {{c_j}} \right\|}}} \right\rangle $, we expect to maximize the cosine proximity between these two vectors, i.e., $\theta = 0$. Hence, $\sum\limits_{i = 1}^n {s_{ij}{u_i}} $ and ${c_j}$ should lie in the same direction, i.e.,

\begin{equation}\label{proof26}
\frac{{c_j}}{{\left\| {c_j} \right\|}} = \frac{{\sum\limits_{i = 1}^n {s_{ij}{u_i}} }}{{\left\| {\sum\limits_{i = 1}^n {s_{ij}{u_i}} } \right\|}}.
\end{equation}

\end{document}